# Collision Avoidance Testing of the Waymo Automated Driving System

Kristofer D. Kusano, Kurt Beatty, Scott Schnelle, Francesca Favarò, Cam Crary, Trent Victor

*Waymo LLC*

## Abstract

This paper describes Waymo's Collision Avoidance Testing (CAT) methodology: a scenario-based testing method that evaluates the safety of the Waymo Driver™ Automated Driving Systems' (ADS) intended functionality in conflict situations initiated by other road users that require urgent evasive maneuvers. The evaluation is performed by comparing the ADS performance to a reference behavior model representing a non-impaired eyes on conflict driver fit to human driver naturalistic data. The Waymo Driver is a SAE Level 4 ADS. The development and deployment of Level 4 ADS is bringing further attention to the challenges associated with the safety evaluation of these novel systems. Unlike advanced driver assistance systems (ADAS), a Level 4 ADS, when engaged, is by definition responsible for the entirety of the Dynamic Driving Task (DDT) execution without reliance on immediate human intervention. This increase in driving role responsibility of a Level 4 ADS leads to a potentially infinite number of operational scenarios in which hazardous situations may unfold. It is thus important to understand what role scenario-based testing can play in the overall safety assurance for level 4 ADS behavioral evaluation, and how to appropriately implement this type of approach, which can be done through a combination of virtual, test-track, and real-world driving environments. To that end, in this paper we first describe the safety test objectives for the CAT methodology, including the collision and serious injury metrics and the reference behavior model used to form an acceptance criterion. Afterward, we introduce the process for identifying potentially hazardous situations from a combination of human crash and near-crash data, ADS testing data, and expert knowledge about the product design and the associated Operational Design Domain (ODD). The test allocation and the test execution strategy are presented next, where test track testing is used to validate the performance estimated in the virtual simulations, which are constructed from sensor data collected on a test track, real-world driving, or from simulated sensor data. The paper concludes with the presentation of results and related discussion from the fully autonomous ride-hailing service that Waymo operates in San Francisco, California and Phoenix, Arizona. CAT is one out of many methodologies used by Waymo to determine safety readiness of the Waymo Driver. The iterative nature of scenario identification, combined with over ten years of experience of on-road testing, results in a scenario database that converges to a representative set of responder role scenarios for a given ODD. When applied in the context of Waymo's virtual test platform, which is calibrated to the data collected as part of many years of ADS development, the CAT methodology provides a robust and scalable safety evaluation solution.

# Introduction

An Automated Driving System (ADS)[1], is designed to perform conditional (level 3) through full (level 5) automation of the driving task as defined in SAE J3016 (SAE 2021). Because ADSs, when engaged, are responsible for the entire dynamic driving task (DDT)[2], considerable effort is needed to verify and validate the safety of an ADS prior to wide deployment on public roads. Scenario-based testing is one method of evaluating an ADS's performance in a set of scenarios (Webb et al., 2020), defined as "a temporal sequence of scene elements, with actions and events of the participating elements occurring within this sequence." (Riedmaier et al. 2020).

Methodologies for identifying the appropriate scenarios to perform a safety evaluation of an ADS are popular research topics, as evident by numerous literature reviews published on the topic (see for example: Riedmaier et al., 2020; Zhang et al., 2021; Ding et al., 2022). A scenario-based approach has also been suggested as a best practice for ADS safety evaluation related to Safety of the Intended Functionality (i.e., SOTIF, ISO 21448:2022) (International Standards Organization [ISO], 2022a) in the standard ISO 34502 (ISO, 2022b). The goal of these scenario-based testing approaches is to perform a "safety evaluation process that identifies risk factors and related potentially critical scenarios that affect the intended functionality, and apply them to evaluate whether ADSs are free of unreasonable risks" (ISO, 2022b, Introduction). This task of identifying a suitable scenario database corresponds to "maximizing the coverage of known hazardous scenarios", as defined in SOTIF clause 7 (ISO 21448, ISO 2022a). Additionally, guidelines being developed by the United Nations (UN) Economic Commissions for Europe (ECE) for Validation Methods for Automated Driving (VMAD) suggest using a scenario-based approach as part of a certification process for an ADS (United Nations Economic Commision for Europe [UN ECE], 2022).

As noted in published literature and standards on scenario-based testing, there are many challenges associated with implementing a scenario-based testing regime for an ADS. First, identifying the appropriate scenarios to evaluate and then demonstrating the coverage of those scenarios is sufficient to accomplish the safety goal is challenging. These first two challenges of identifying scenarios and demonstrating coverage have been addressed by various entities for SAE level 3 systems focusing on an operational design domain (ODD)[3] covering highway driving automation in Germany (German Aerospace Center, 2019) and Japan (Antona-Makoshi et al., 2019; Nakamura et al., 2022). The challenges related to defining the scenario space become even greater when considering an SAE level 4 system that operates in a dense urban environment because of the complexity introduced by the increased number of possible maneuvers and interactions that need to be considered. To date, details about large-scale scenario-based testing programs for a level 4 ADS have not been examined to the same extent as level 3 programs. The Safety Pool Initiative is developing a scenario "database of curated driving scenarios shared across stakeholders and geographies, where organizations can exchange, test, benchmark scenarios and use the insights to inform the making of policy and regulatory guidelines" (Safety Pool, n.d.). The PEGASUS family of projects, through the Verification & Validation Methodology (VVM) Project (Verification and Validation Methods, n.d.) and SET Level Project (Set

---

[1] An ADS is defined as "The hardware and software that are collectively capable of performing the entire DDT on a sustained basis, regardless of whether it is limited to a specific operational design domain (ODD); this term is used specifically to describe a Level 3, 4, or 5 driving automation system." (SAE 2021).

[2] The DDT is "all of the real-time operational and tactical functions required to operate a vehicle in on-road traffic, excluding the strategic functions such as trip scheduling and selection of destinations and waypoints, and including, without limitation, the following subtasks: 1. Lateral vehicle motion control via steering (operational). 2. Longitudinal vehicle motion control via acceleration and deceleration (operational). 3. Monitoring the driving environment via object and event detection, recognition, classification, and response preparation (operational and tactical). 4. Object and event response execution (operational and tactical). 5. Maneuver planning (tactical). 6. Enhancing conspicuity via lighting, sounding the horn, signaling, gesturing, etc. (tactical)" (SAE 2021)

[3] ODD is defined as "operating conditions under which a given driving automation system or feature thereof is specifically designed to function, including, but not limited to, environmental, geographical, and time-of-day restrictions, and/or the requisite presence or absence of certain traffic or roadway characteristics." (SAE 2021)

Level, n.d.) are developing simulation-based methodologies to evaluate ADS safety. These methodologies and toolchains provide guidance on how to define critical scenarios for evaluations and the simulation toolchains to do virtual testing, among other topics. Even with situations defined, because of the so-called "parameter explosion" problem of having too many combinations of distinct scenarios and parameters to reasonably evaluate, determining which scenarios to evaluate is also an additional challenge (IEEE 2846-2022, Annex A). Due to the greater complexity of urban driving, the "parameter explosion" problem is even more challenging than in highway driving.

The goal of this paper is to describe the Collision Avoidance Testing (CAT) methodology used to evaluate the performance of the Waymo Driver™ ADS in urgent situations where the ADS responds to some surprising actions of another actor, i.e., the responder role[4] (Scanlon et al., 2021; Scanlon et al., 2022) in order to avoid or mitigate collisions that are relevant to a given ODD. The CAT methodology is a scenario-based framework that performs a prospective evaluation of the ADS performance prior to starting to operate the ADS without a human driver behind the wheel in scenarios that might reasonably be expected to be encountered, even if rarely, during public road testing during development. As discussed in Webb et al. (2020), the CAT methodology is one of many complementary methods used to determine the readiness to start fully autonomous operations, that is, an SAE Level 4 system operating without a human driver, either in the driver's seat or remotely. Examples of fully autonomous operations are the Waymo One™ commercial service in Arizona, where members of the public can hail a fully autonomous passenger vehicle without a human driver in the area in and around Chandler, AZ. This paper will present the components and process (safety test objectives, scenario identification and selection, and the test execution) used for the CAT methodology and how Waymo applied these methods to start providing fully autonomous ride hailing trips in San Francisco, California and Phoenix, Arizona. Today, this ride hailing ODD consists of non-limited access roads and excludes parking lots in densely populated urban areas but is subject to expansion to include limited access highways in the future.

## Methodology

### Methodology Overview

Figure 1 shows an overview of the process elements necessary for execution of the CAT methodology, each of which are discussed in more detail in the following sections. The scope of the CAT methodology is to evaluate the performance of the ADS in scenarios where the ADS uses urgent evasive maneuvers to respond to the surprising actions of another actor to avoid or mitigate. These types of responder-role situations are important to include in a safety evaluation of an ADS because ADSs will continue to operate in a mixed traffic environment for most proposed use cases and will likely encounter some of these traffic participants performing unexpected, or surprising, maneuvers (Scanlon et al., 2021). Thus, although the set of scenarios envisioned in an AV-only traffic environment may be somewhat different, ADS will continue to encounter these types of scenarios experienced by today's human drivers. Further, these maneuvers may include routine driving interactions that are experienced by drivers frequently, such as other actors suddenly changing lanes. The maneuvers, however, could also include more extreme behaviors that are rarely encountered during routine driving, such as drivers running red lights midphase at high speeds. Responder role collision avoidance is especially challenging in a typical ride hailing ODD, which features densely populated roadways with complex traffic patterns where vehicles and vulnerable road users often share the trafficway. These interactions with vulnerable road users are particularly important because of an increased injury risk given a collision compared to vehicle occupants.

In the CAT methodology, the ADS is evaluated under key identified collision avoidance scenarios, where ADS failure situations (e.g., sensor failures) are not evaluated. These failure situations are evaluated and mitigated

---

[4] The responder role is the actor in a conflict that "must take some action in response to the series of events caused by the initiator party's initial, surprising action." (Scanlon et al., 2021)

through other methodologies such as hazard analysis, platform verification and validation (V&V), and simulated deployments (Webb et al. 2020). Waymo is currently developing the Waymo ADS for multiple applications (e.g., transport of people in a ride-hailing service, and transport of goods in a delivery and freight transportation service), that operate in different ODDs (e.g., urban road configurations and traffic conditions in Phoenix, AZ and San Francisco, CA for ride-hailing and highway road configurations and speeds between Houston and Dallas, Texas for freight transportation) and vehicle platforms (i.e., passenger cars and heavy commercial vehicles). The CAT methodology, therefore, was designed to be adaptable as different applications of the Waymo ADS are prepared. The CAT methodology starts in the early stages of the product design cycle before fully autonomous operations and continues through the in-field operation of a system.

The safety test objective is to demonstrate that the ADS meets or exceeds the level of performance of a reference behavior model based on data from human drivers that are **N**on-**I**mpaired and **E**yes **ON** the conflict (NIEON) (Engström et al., 2022; Scanlon et al., 2022) . The acceptance criterion and model are presented in more detail in the "Safety Test Objective" section below.

The CAT methodology needs to balance several priorities in the evaluation process: test execution time, completeness of execution, and fidelity of the results. Waymo uses an incremental approach to fully autonomous operations that contributes to the readiness determination decision that expands the ODD as evidence of proficient performance is accumulated. As such, the ODD may start in a small geographic area and limited operating conditions then be expanded a short time after (e.g., within months) to include more areas and more challenging conditions for performance of the DDT. The results of the CAT methodology are one of the complementary methods used leading up to the start of fully autonomous operations to help gauge readiness and to plan for future ODD expansion. The CAT evaluation needs to be completed in a timeframe of several weeks after a software candidate is identified. To meet this goal, the CAT approach relies heavily on virtual testing to complete the large number of scenarios in such a timeframe. At the same time, to have confidence in the results, a high level of simulation fidelity, especially in the perception system, is an important part of the virtual test platform. To accomplish both the goals of reasonable test execution time and high level of simulation fidelity, the CAT methodology has primarily used sensor data collected on a test track to construct virtual simulations that can be evaluated on new software releases. As sensor simulation technology has progressed over the past few years, the share of synthetic scenarios, that is, those where the perception sensor data is generated from a model of the actual sensor performance, has increased, especially in scenarios where it is challenging to collect test data on a test track. Using actual or simulated sensor data allows a realistic representation of the perception performance including noise, which can influence system performance in these types of urgent, responder role scenarios. The test allocation and the capabilities of the virtual test platform are described in more detail in the "Test Case Allocation" section below.

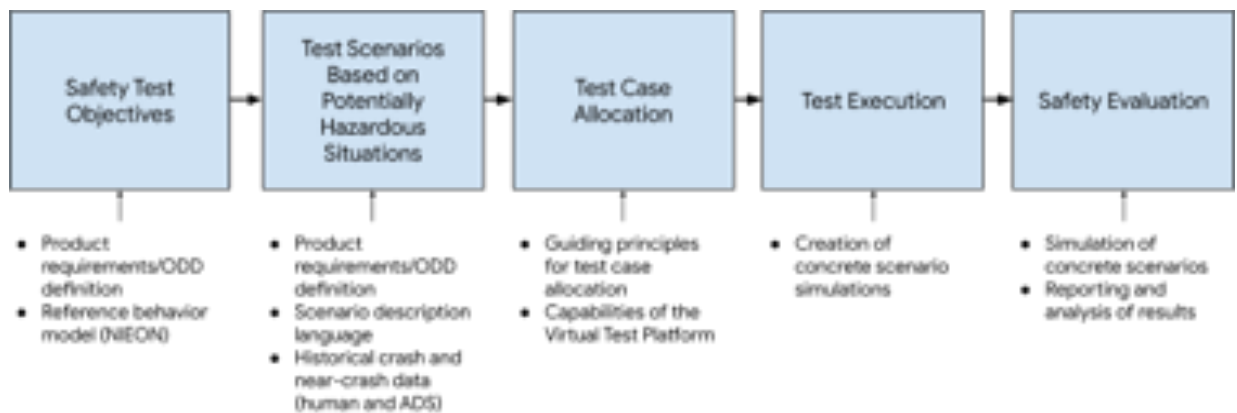

**Figure 1.** Visualization of Organization of Scenarios for the Collision Avoidance Testing (CAT) Safety Evaluation.

# Safety Test Objective

## *Method to Set the Test Objective*

Safety objectives represent the sought safety properties that a developer would like to demonstrate through a particular testing campaign (ISO 34502:2022). Waymo's approach for the determination of readiness of the Waymo Driver rests on the need to demonstrate that the ADS does not pose an unreasonable risk to safety of the public (Webb et al., 2020). The ISO draft standard on scenario-based safety evaluation (ISO, 2022b) brings forth two possible ways to approach the setting of safety test objectives: (a) the specification of "an upper boundary value of the acceptable occurrence rate of a measurable behavior of the ADS" (e.g., "hazardous behavior of the system shall occur less than once per x hours during operation within the operational domain"); and/or (b) the specification of "a performance reference model regarding the capability of the ADS to handle certain scenarios safely, based on minimum performance levels required for these scenarios" (e.g., "The ADS shall be capable of preventing any accident that would be preventable according to a reference performance model of a competent and careful human driver"). The two approaches are not in opposition with each other. In fact, one could argue that approach (a) allows us to reason about the aggregate-level performance on a statistical basis, which is necessary when the determination of acceptability of overall residual risk has to be done, while approach (b) enables the event-level reasoning that ensures no single scenario can unreasonably contribute to such residual risk.

Waymo combines both approaches to establish the acceptance criterion for the CAT methodology. Specifically, the acceptance criterion for the CAT methodology is predicated upon a comparison between the Waymo ADS and that of a behavioral reference model termed "NIEON" (an approach akin to option (b)), where we require comparable or better performance across appropriately set scenario safety groups and road users groups (thus setting targets with an approach akin to option (a)).

An approach that combines the identified options rather than picking one is in line with the multi-pronged approach to safety determination presented in Webb et al. (2020). The CAT methodology, thus, recognizes the advantages of combining both approaches proposed in ISO 34502:2022 to help ensure that satisfactory performance at an aggregate level does not unintentionally conceal concerning performance at single scenario levels. At the same time, however, it is not practical or feasible to outperform the behavioral reference model in every single scenario variation. For example, the implemented reference model is evaluated in both swerving left and right in addition to only braking. The best outcome out of the three possibilities is chosen as the NIEON reference. Humans need to decide which direction to swerve and may choose one direction over the other in certain scenarios. The current NIEON reference model benefits from the best outcome (without having to choose, as a human would). The concerns related to having acceptable performance at higher levels of aggregation (e.g., all collision scenarios) yet deficient performance in lower levels of aggregations (e.g., in specific collision modes or with certain actor types) coupled with the impracticality of testing millions of scenario variations can then be mitigated by aggregating lower levels of safety-relevant scenarios and identifying minimum benchmarks of reasonable performance for these groups.[5] Moreover, aggregating at these lower levels of related driving scenarios that are ODD-relevant provides benchmarks that are better sourced and more indicative of actual ADS performance than a single target aggregated over all scenarios, which is insufficient for demonstrating ADS safety.

The effectiveness of the approach would thus rely, in turn on: 1) the appropriate selection and implementation of the reference behavior model to be employed in the analysis (in our case, the NIEON model); and 2) the appropriate definition of lower levels of aggregation for scenario groups and the selection of the tested scenario variations in the first place.

---

[5] The tension between aggregate-level approaches and those requiring appropriate reasonableness of performance at the event-level was also portrayed by Favarò (2021), who recognized that different stakeholders may be interpreting aggregate approaches' value and sufficiency in different ways.

The selected reference behavior model (the **N**on-**I**mpaired and **E**yes **ON** the conflict model driver) represents a level of performance that does not exist in the current human driving population, as all human drivers engage in activities that take their gaze or attention off the roadway for some amount of time (e.g., adjusting vehicle controls, reaching for objects). The implementation of the NIEON model response time (Engström et al., 2022) and avoidance maneuver (Scanlon et al., 2022) is based on state of the art human behavior research and available human data. The notion of NIEON as a reasonable benchmark for high quality driving is supported by the results presented in Scanlon et al. (2022), where the NIEON reference model was able to prevent 84% of serious injury risk in the responder role of scenarios reconstructed from real-world collisions that resulted in a fatality. Still, like any model, NIEON is subject to simplifying modeling assumptions, limitations, and variability in its application (e.g., parts of the model implementation and/or application to certain scenarios requiring expert judgment). We believe the NIEON model is a sufficient benchmark for collision avoidance behavior because it represents the definition of reasonable human behavior when driving, that is to be attentive and non-impaired, in addition to the result that suggests the NIEON model can prevent or mitigate a large proportion of fatal collisions.

With regards to the creation of an appropriate library of scenarios and the definition of relevant scenario groupings, our focus is on the process to identify and continuously add scenarios that are representative of the types of responder role scenarios we reasonably might encounter in a given ODD. Scenarios in the CAT library do not have an associated weighting function that ties the scenario to the probability of occurrence in a given ODD. The goal is thus not that of estimating an overall crash rate and probability of occurrence from the testing campaign, but rather to help ensure that the overall set of scenarios is representative of the behavioral competencies incorporated into the ADS that are relevant to safe responsive performance in the ODD's conditions. The CAT program thus provides support, in conjunction with other methodologies, that the intended system does not present an unreasonable risk to motor vehicle safety, which is aligned with the goals of the SOTIF standard (ISO, 2022a). Further details on the scenario groups definitions and the scenario identification are presented in the next sections.

## *Metrics to Measure the Test Objectives*

To assess the ADS's performance relative to the NIEON model, the CAT methodology uses an aggregate scenario scoring mechanism. The aggregate scoring is performed across groupings of scenarios, as shown in Figure 2. In Figure 2, the possible scenario space is represented as a 2-dimensional space. Scenarios that are applicable to a given ODD are grouped into distinct safety groups, which are defined by conflict type and conflict partner. Each safety group can contain multiple sub-categories of scenarios. For example, a safety group of "Vehicle Cutting Across, Perpendicular" includes subcategories of "left turn across path", "right turn across path", and "straight crossing path" which each can contain multiple individual scenarios. Additionally, the scenarios are aggregated into two high-level road user groups of Vehicles and vulnerable road users (VRU, that is, pedestrians and cyclists). The aggregation is performed within the road user groups and safety groups to ensure that the Waymo ADS provides sufficient confidence that the ADS meets a level of acceptable risk across all expected responder role scenarios, without compromising performance in one area for performance in another. The acceptance criterion for the CAT methodology is predicated upon a comparison between the Waymo ADS and the NIEON model performance, where we target comparable or better performance within each safety group and for the two high-level road user groups of Figure X. The performance comparison is performed based on the number of total collisions (contact between the ADS and another actor) and serious injury events, where the same event severity model (discussed below) is employed to assess the Waymo ADS and the NIEON model performance. The acceptance criterion is that, within safety groups and for the two high level road user groups, the Waymo ADS shall have as good or better performance compared to the reference behavior model in both the number of collisions (contact between actors) and serious injury events (based on an event severity model discussed below). The determination of the safety groups and the scenarios to consider is discussed in more detail in the following section.

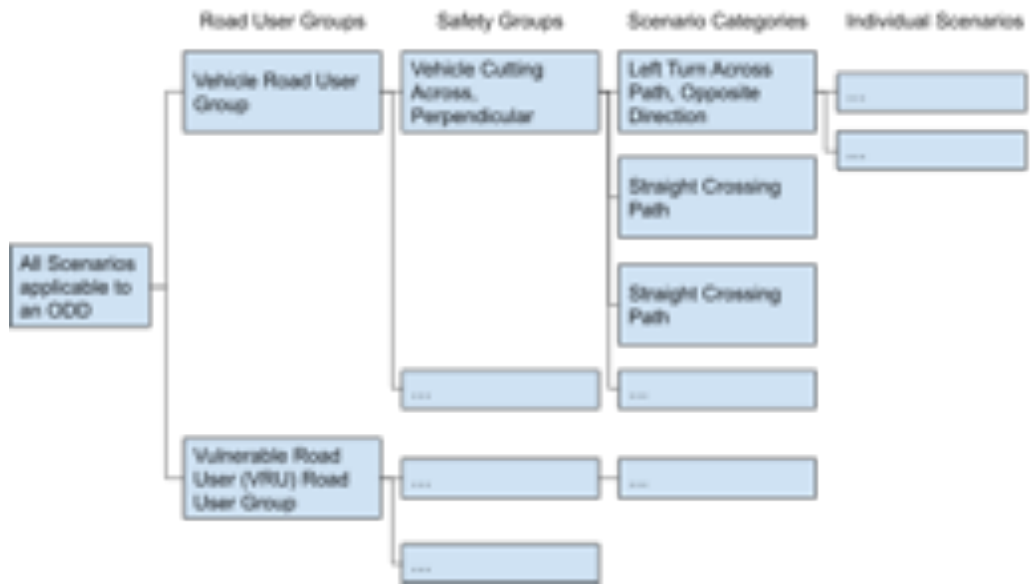

**Figure 2.** Example Visualization of the Organization of Scenarios for the Waymo CAT Safety Evaluation.

There are two metrics considered for the safety test objectives, one related to collisions and one related to injury-causing collisions. A collision is defined as contact between the ADS vehicle and another actor during the scenario. The collision metric excludes collisions that involve impact exclusively to the rear ⅔ of the ADS vehicle (i.e., non-frontal) and collisions that occur while the ADS vehicle is stationary. The frontal collision requirement was put into place to make it easier for evaluators using this collision metric to identify collisions where the ADS behavior, even though responsive to surprising behavior by other road users, contributes to the collision outcome. This frontal collision assumption does not affect the safety evaluation results because the ADS and reference model would have the same collision outcome in such cases even if the requirement was not used, yet it has the technical benefit of making the simulation and interpretation of the results more efficient. The test objective of the CAT methodology is to demonstrate the ADS has comparable or better collision avoidance behavior compared to the NIEON model in responding to conflicts initiated by others. As a result of the ADSs' collision avoidance maneuvers to respond to the surprising event, the ADS may come to a stop or enter adjacent lanes of travel, which could result in other road actors not involved in the original conflict to collide with the ADV. Estimating the risk associated with these potential subsequent conflicts is not within the scope of the CAT methodology at this time as it would become a probabilistic estimation of other actor behavior.

The injury-causing collision metric uses the characteristics of the impact dynamics and injury risk models to estimate a probability of an injury occurring. These models are the same impulse-momentum based collision model and omni-directional injury risk model used for vehicle-to-vehicle collisions (McMurry et al., 2021) and impact speed based risk functions for pedestrians, cyclists, and motorcyclists as described in Scanlon et al. (2022). The output of the injury risk model is a probability of Maximum Abbreviated Injury Score (MAIS) of level 3 or greater (MAIS3+), i.e., p(MAIS3+), for each actor involved in the collision. The injury risk is computed for the ADS vehicle and the NIEON reference model, if a collision occurred. A serious injury event is defined as a threshold on the predicted p(MAIS3+) from the collision severity model. A p(MAIS3+) $\geq$ 10% is used in ISO 26262 as the threshold for a Severity level 2 (S2) event. We chose thresholds of p(MAIS3+) $\geq$ 5% for vehicle-to-vehicle collisions, p(MAIS3+) $\geq$ 1.5% for children pedestrians, and p(MAIS3+) $\geq$ 10% for all other pedestrian, cyclist, and motorcyclist actors. Because the injury risk curves used for pedestrians did not specifically account for the pedestrian age, a lower threshold for children was chosen to account for increased risk due to higher likelihood of engagement of children with the lower portion of the vehicle. The injury-causing collision metric does not use the non-frontal collision exclusions used in the collision metric discussed previously.

Both metrics employed in the CAT methodology (i.e., the collision metric and the injury-causing collision metric) inform the evaluation of the acceptance criterion established for scenario-based responder-only collision avoidance testing. As mentioned in the beginning of this section, the acceptance criterion for the CAT methodology rests on a comparison between the Waymo ADS collision avoidance performance and that of an NIEON reference behavior model. The stringency of the criterion can be set by modifying the modeling parameters associated with the NIEON model. At present, the CAT methodology does so by appropriately selecting the specific response time within which the NIEON agent acts.[6] The response time model employed was presented in Engström et al. (2022). That paper describes how the NIEON response time modeling approach fits a statistical distribution to human drivers with their eyes on the road at the time of the conflict. Specifically, the model features a linear fit of response time as a function of ramp-up time. This ramp-up time is the difference between the stimulus onset (i.e., the start of the surprising event) and stimulus end. The shorter this ramp-up time, the faster the response time. The linear relationship can thus be identified through the specific slope and intercept. Modifying the intercept of the response time model produces a function that is conceptually similar to selecting different percentiles of the human population. Waymo adopts a different benchmark for the Vehicle and VRU road user groups: in particular, the Vehicle road user and safety groups use a larger intercept than the VRU road user and safety groups. This implies that the comparison point used for VRU groups is more stringent than that used for Vehicle groups. Such a decision is grounded in Waymo's use-case and location of operations, which include dense urban environments, and was informed by the distribution and likelihood of potential high-severity outcomes associated with VRU interactions. Furthermore, the selection of intercept and associated stringency of the NIEON model is, for all groups, grounded in the establishment of a minimum level of acceptable risk that is necessary to ensure absence of unreasonable risk. Waymo employs a combination of acceptance criteria associated with multiple methodologies that allow Waymo to validate the Waymo Driver performance (Webb et al., 2020). As part of this Waymo's risk management approach, Waymo also ensures that targets for both Vehicle and VRU groups are continuously revised and made more restrictive, toward goalpost targets that Waymo sets for full-scale deployment and that enable our company philosophy of reducing traffic injuries and fatalities a reality at-scale.

## Test Scenarios based on Potentially Critical Situations

At the outset of development, test scenarios in the CAT methodology are developed systematically, considering the existing scenario-based tests from the many years of Waymo's ADS development, the ODD, system capability, and human crash data (e.g., police accident databases, naturalistic driving studies). As development and testing progresses, applicable conflicts identified from on-road testing or simulations are used to confirm coverage of considered test scenarios and to add coverage if novel scenarios are identified. Figure 3 summarizes the test scenario identification process. This process identified in Figure 3 has a similar intent to section 4.3 of ISO 34502 (ISO, 2022b). The following paragraphs describe each of the processes and data sources that are used to determine the set of potentially relevant situations.

---

[6] At this time Waymo does not leverage variation of the avoidance maneuver model (i.e., the modeling of steering and braking actions).

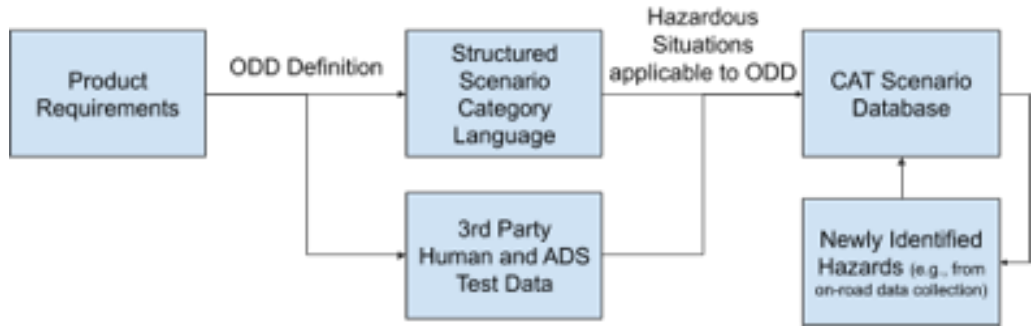

**Figure 3.** Process for Determining Test Scenarios based on Potentially Critical Situations for the CAT Methodology.

First, a high-level scenario description language was developed that considers actor types (e.g., passenger vehicle, heavy vehicle, cyclist, pedestrian) and starting and ending location (e.g., within lane, across lane, off road), traffic control devices (stop signs, traffic signals, etc.), roadway features (e.g., shared turn lanes) and maneuver type (e.g., turn left). This scenario description language aids in performing a structured, combinatorial analysis of potential methods, similar to the method in Annex E (Derivation and Structuring of Scenarios using Criticality Analysis) of ISO 34502 (ISO, 2022b) or other structured, scenario generation process like presented by Thorn et al. (2018). The methods for selecting representative scenarios to meet the goals of having a tractable number of scenarios to evaluate is discussed in more detail in the [Test Case Allocation](Test Case Allocation) section below. At the start of development, the ODD as described in the product requirements is compared with combinations of the scenario description language to determine which interactions between agents are applicable. For example, if an ODD excludes unprotected left turns, then scenarios with the ADS vehicle turning left at unprotected intersections need not be considered. Additionally, combinations of scenario descriptions that would not result in a conflict are excluded. For example, another actor traveling in the opposite direction as the ADS and taking a right turn while the ADS takes a right turn could not result in a conflict, and thus is not considered for the scenario database. In addition to the core scenario descriptions that describe the geometry of a conflict, a number of salient factors were identified which are properties of the actors or scene that may impact the performance of the system in material ways. Examples of salient factors are objects that are occluded, whether the scenario occurs on a high grade (slope), and lighting conditions (e.g., low sun angle). These salient factors are also considered in making the high-level scenario descriptions.

Second, available human collision data sources are cross referenced to confirm that the combinations of the scenario description language cover all known crash modes. This comparison to human data is especially important for the scope of responder role evaluation. We assume that, as the ADS operates in a mixed traffic setting, the human actors will continue to initiate similar conflicts that ADS has to respond to. We considered nationally representative crash databases (e.g., the National Highway Traffic Safety Administration's (NHTSA) Fatal Accident Reporting System (FARS), the Crash Report Sampling System (CRSS), and Crash Investigation Sampling System (CISS)). Because these crash data sources are either based on police accident report data or retrospective crash investigations, they lack detailed information about the pre-crash conditions and movements of actors that may affect crash avoidance performance. Additionally, crash databases are known to have underreporting of certain types of collisions. For these reasons, we also consider naturalistic driving study (NDS) data that include video and additional sensors from human crash and near-crashes, such as the Strategic Highway Research Project 2 (SHRP-2) Naturalistic Driving Study (Hanky et al., 2016) and other proprietary sources (e.g., dash cams). An additional use of the crash and naturalistic data is to parameterize the behaviors of actors (e.g., travel speeds, accelerations). We also utilize data collected from our own vehicles, which record the behavior of surrounding traffic.

The scenario description language analysis, human crash databases, and ADS functional design are the primary sources for the initial development of the considered scenarios, especially during the early parts of the development phase. For subsequent applications of the ADS (e.g., when expanding to a new ODD, the considered scenarios from previous ODDs serve as the starting point for the new scenario database. In the case of a subsequent application of the ADS, the CAT methodology uses a structured salient factor analysis to determine what new test coverage should be added based on unique characteristics of the ODD that differ from the past ODDs. The end goal of this salient factor analysis is to (a) determine which scenarios from previous ODDs apply to the current ODD, (b) which new scenarios not considered in previous ODD are now applicable and need additional coverage. These salient factors of a new ODD are determined from the product requirements, known system performance, and analysis of the current traffic situation in the new ODD.

Last, as development on an ADS application matures and accumulates more data gathered from test vehicles in the ODD, there is a feedback loop in CAT methodology that confirms that events experienced during on-road testing (either simulated, such as after an autonomous specialist, or safety test driver, disengagement, or with the ADS in control, see Schwall et al., 2020) are sufficiently covered in the scenario database. Relevant conflicts (i.e., responder role scenarios) are first identified from on-road testing data and then compared to the scenarios already identified. If the scenario from the real-world data is not present, additional scenarios are added by creating a set of scenarios that represent the observed real-world event. Additionally, even if the observed real-world scenario is already considered in the scenario database, the real-world example itself or a set of scenarios representing that event could be added to increase coverage. In addition to the ADS test data, new sources of human crash data may become available. For example, crash reconstructions of fatal collisions in one ODD in Chandler, AZ were performed as part of Scanlon et al. (2021), which was separate from the CAT methodology. As these reconstructions were performed, the scenarios were compared with the scenarios in the existing scenario database.

## Test Case Allocation

### *Guiding Principles for Test Case Allocation*

Several guiding principles are applied when considering test case allocation for the CAT methodology to meet the goals of timely test execution and a high fidelity result.

First, because the acceptance criteria for the safety test is a comparison of the ADS performance to a reference behavior model (the NIEON model), the scenarios selected for testing must be able to differentiate between passing and failing behavior of the reference model. That is, scenarios in which the reference model easily passes (are "too easy") or has no chance of avoiding ("unavoidable") need not be oversampled to achieve the test objective. This sampling approach is chosen given the existence of other methodologies used in Waymos readiness determinations, such as hazard analysis, platform V&V, and simulated deployments (Webb et al., 2020) that account for the potential of novel ADS-failure that may lead to collisions even in situations that the reference model could easily avoid. Furthermore, considering that the CAT method employs a scoring methodology that compares the ADS and reference model performance, oversampling of the "too easy" or "too hard" space would not affect the results as both the ADS and reference model would have the same outcome. Oversampling of scenarios that are unavoidable by the reference model has also limited utility, considering how such a reference model was established as a benchmark for reasonable preventability based on a NIEON human driver, as discussed in the Safety Test Objective section. To determine the appropriate mix of scenarios, the reference behavior model response time is varied by several fixed intervals. A single-proportion z-test is then applied to the rate of failure of the different reference behavior model levels to determine if the rates are statistically different. This z-test is performed on the safety groups that make up the aggregate scoring. Additional scenarios are collected for safety groups that do not have a statistically significant difference in passing rate collision metric. This procedure ensures the evaluated concrete scenarios can be used to reliably differentiate passing and failing behavior of the reference model, and thus helps ensure an appropriate benchmark for the ADS evaluation. Note that this statistical diagnostic is performed only on the NIEON model results and is independent of the ADS performance. The statistical test helps ensure that the scenarios in the scenario database can differentiate between different performance levels of the NIEON model, and thus by extension can be effectively used to compare the NIEON performance to the ADS to accomplish the safety test objectives.

Second, the test case allocation used in the CAT methodology favors a segmentation approach to determine representative scenarios for all hazardous scenarios over an approach that exhaustively samples the logical scenario space. Many of the existing scenario-based testing methodologies discussed in the introduction of this paper are scoped for highway driving and some further scoped to low-speed (i.e., traffic jam assist) applications. Given the large number of scenarios that need to be considered in the open context of driving, an exhaustive sampling of all parameter types (e.g., all colors of vehicles) would cause the number of scenarios to become excessively large, which is often referred to as the "parameter explosion" problem. The segmentation approach is suggested in ISO 34502 (ISO, 2022b) as a possible solution to the parameter explosion issue. Given the requirements of reasonable test set execution time discussed in the Methodology Overview section, the CAT methodology prioritizes breadth of coverage instead of exhaustive sampling of parameter ranges and their combinations. As discussed above, this segmentation approach was chosen in the context of other methodologies utilized by Waymo for determining readiness (Webb et al., 2020). The type of "unknown-unknown" scenario discovery that may be achieved by exhaustive or adaptive scenario generation are achieved through other simulation efforts (such as simulated deployments), hazard analysis, and platform V&V. The concrete scenario selection process is discussed in more detail in the following "[Test Execution](#)" section.

Third, a high level of simulation fidelity in the virtual test platform is required to accurately evaluate the ADS performance. Some scenario-based testing methods used in the past have greatly simplified simulation environments that either ignore perception performance, or model things like perception performance in simplistic ways like applying a correction factor for overall system performance (ISO, 2021). Given an ADS will encounter a broad variety of driving situations, perception system performance is an especially important consideration when evaluating an ADS performance. However, given the execution time requirements a large number of test track tests cannot be feasibly completed for each software release before starting fully autonomous operations. To meet the requirements of reasonable test set execution time and a high fidelity representation of the perception system performance, the CAT methodology utilizes several scenario creation methodologies. The first and most common approach used is to collect sensor data from an instrumented vehicle on a closed test track and create a scenario from this sensor data that can be used to simulate the system performance on new software releases. Additionally, when the scenario is created from this logged data, the positions or speeds of other actors can be modified to make the scenario more likely to require urgent evasive maneuvers to avoid or mitigate a collision. This allows for the test drivers collecting the data to use real vehicles interacting in a safe manner while still being able to construct relevant scenarios. Surrogate test devices, such as pedestrian mannequins, are also used in test track testing when it is deemed too hazardous to create a scenario with human participants. Similarly, if a hazardous situation (e.g., a novel scenario where a specific type of sensor data may have contributed to the difficulty of the scenario) is discovered while performing on-road testing, the same procedure as is used for the test track data can be used to add the on-road testing data as a scenario in the database. Other scenarios, however, are either too dangerous (e.g., at high speeds) or impractical (e.g., those requiring specific intersection geometry or vehicle types such as light rail not available on a test track) to collect on a test track. For these scenarios, fully synthetic simulations are created. These fully synthetic simulations use sensor simulation that attempts to recreate the range, field of view, and noise characteristics of the actual sensors.

All scenarios used for the safety evaluation of a potential software release, whether derived from real-world data, test track data, or synthetic means, are evaluated in virtual simulation. The virtual test platform's ability to reproduce the same outcomes as the software under test on a test track are used as validation of the test platform's capabilities, however, as discussed in the section below. Waymo also uses other methodologies, such as simulated deployments, to estimate the performance of the ADS based on data collected from public road testing (Webb et al., 2020). These simulated deployments use previously recorded driving data to simulate the operation of the ADS and includes a range of possible interactions, not only collision avoidance situations.

## Safety for Data Collection

Although the safety evaluation in the CAT methodology is exclusively done using a virtual test platform, much of the underlying sensor data used in these virtual simulations are collected from the real world on a test track. As these types of responder role scenarios feature situations requiring urgent evasive maneuvers to avoid a collision, special care is taken to ensure safety during data collection on a test track. Waymo employs a standardized procedure for test track scenario data collection that helps ensure safety. For each scenario, there are several defined roles for those who collect the data, including one or more drivers, co-drivers, participants, and test coordinators. Test requests are submitted that describe the desired scenario to collect, which includes diagrams of the actor movements and descriptions of the environment and parameters (e.g., speeds, angles, etc.). The test coordinator then assigns personnel roles according to the test plan. Prior to testing, the test team meets to review the plan for the testing. Next, the test personnel perform a series of "dry run" events, where only one actor in the scene is moving at a time. The purpose of the dry run events is to have the test personnel become familiar with the required maneuvers of the test. The dry runs are also used to identify and mark maximum braking points and do-not-cross lines. A potential conflict point, the point where the responding vehicle (i.e., the ADS vehicle) needs to apply the brakes in order to avoid reaching the conflict point, is identified. The end of this braking zone is the do-not-cross line, which indicates the area where the other participants should avoid entering. The dry runs are also used to practice the exit strategies for the participants to leave the conflict point after the scenario is executed. Test personnel are trained for each specific role using a combination of classroom and hands-on instruction. This training is refreshed periodically with safety assessments and re-training.

The test procedures process was developed by using a system theoretic process assessment (STPA) (Leveson & Thomas, 2018). STPA is a top-down, structured analysis method that identifies unsafe control actions that may lead to hazards. The process analyzes potential unsafe control actions for each test participant and each step in the test process (start, conducting, ending). The output from this STPA analysis is a tree-like hierarchical structure that can help develop safety requirements and mitigations. Based on this analysis, the test collection procedure was developed to include sufficient mitigations for all of the potential unsafe control actions. An example of a mitigation implemented from this analysis is having a secondary test coordinator to review the test plan to ensure it complies with all policies.

## Capabilities of the Virtual Test Platform

Because the CAT methodology exclusively uses a virtual test platform for test execution, the success of the safety evaluation is heavily dependent on the capabilities of the virtual test platform. Over many years of developing and deploying ADSs, Waymo has developed a large distributed software in the loop (SiL) simulation platform that is used for the CAT evaluation. The simulation environment models the vehicle dynamics and control systems of ADS and executes the vehicle behavior software modules responsible for perception, planning, and behavior prediction used in the on-road vehicle. As discussed above, collected sensor data from test tracks or simulated sensor data is used in these SiL simulations. In the system that runs on the vehicle, individual software modules run independently and asynchronously. The SiL simulation environment does not run in real time. If the software modules were allowed to run with no constraints, they would likely be able to complete their computations before new sensor data is simulated. To accurately model system latency introduced by these different module execution times in the simulation environment, the 95th percentile in-field execution time of each module is applied to each software module. The SiL simulation platform is the same as the one used for the study on ADS performance in reconstructed fatal collisions by Scanlon et al. (2021, 2022). These studies implemented the simulations using synthetic sensor simulation as opposed to using test track collected sensor data, as used in the CAT methodology for safety readiness determinations.

The virtual test platform produces repeatable results even though the simulation platform has the possibility for non-determinism. For example, inter-module communication is not guaranteed to be deterministic, so the order in which each module receives information may be different from run to run. The effect of this non-determinism on the current evaluation is small. The CAT safety evaluation was performed approximately 10 times with the same software and the change in the aggregate score as a percentage of the number of scenarios in each safety group was evaluated. This repeated evaluation found that the 95th percentile of this percent change in the aggregate score was 1.5% for the collision metric and 0.6% for the serious injury event metric.

The virtual test platform provides a conservative estimate of the number of simulated ADS collisions when compared to the same scenarios executed on a test track. For each start of fully autonomous operation, a series of approximately 70 scenarios are reproduced on a test track and executed using the same candidate software as is being considered in simulation. These scenarios involve soft test mannequins that are utilized in other active safety test track evaluations that minimize the hazards to the test performers if a collision should occur. The scenarios are executed with an Autonomous Specialist behind the wheel of the ADS where the ADS will continue its test in autonomous mode even if a collision is to occur. As a result, this type of testing is only feasible for scenarios that utilize soft targets, such as pedestrian and cyclists mannequins, that can be struck by the ADS vehicle without harm to participants or major test vehicle damage. The verification criteria for the reproducibility of these tests are: (a) they simulations provide a conservative estimate of the number of collisions observed on the test track (i.e., there are as many or more simulated collisions than test track collisions) and (b) the simulated ADS trajectories are ahead of the test track trajectories (i.e., the position of the simulated vehicle is even with or ahead of the test track trajectory at the same reference time) for a majority of simulations.

Like in many other virtual test platform validation approaches (e.g., ISO/TR 21934 [ISO, 2021], ISO 34502 [ISO, 2022b], and UN ECE, 2022, Annex III), the simulation environment should be as accurate as necessary to evaluate the safety test objectives. Hardware- and/or Vehicle-In-the-Loop (HiL and ViL, respectively) approaches have been suggested for use in a virtual test platform. Waymo does use HiL and ViL in other complementary readiness determination methods, like the platform V&V program, to verify that the hardware systems of the vehicle meet their designated requirements. These approaches would, however, greatly increase the cost, complexity, and time necessary to perform the CAT evaluation, which would not meet the test objectives of repeated, short cycle-time evaluation laid out in the "guiding principles for test allocation" section. Additionally, although using HiL and ViL would increase the accuracy of reproducing the computing and networking performance of an actual vehicle platform, they would not completely address other important aspects of the fidelity of the virtual test platform, such as realistic sensor data. The validation approaches laid out in the previous paragraphs, in addition to the experience in building the simulation platform to reproduce the many years of on-road testing that Waymo has conducted, demonstrate that the virtual test platform provides a credible estimate of the satisfying the safety test objectives.

## Test Execution

A segmentation approach is utilized in the CAT methodology to select concrete scenarios that are representative of the hazardous situations identified for an ODD. First, a combination of the high level scenario description language is considered, as described in the previous "[Test Scenarios based on Potentially Hazardous Situations](#)". This includes a description of the high level road features, actors, maneuvers, and geometries of the interaction. Next, that high level scenario description is decomposed into relevant parameters and salient factors. At this stage, several potential scenario features can be combined to create representative scenarios. For example, if the high level scenario is a passenger vehicle pulling out from a driveway into the path of the ADS, a potentially representative scenario involves the ADS traveling in the rightmost lane, closest to where the vehicle will pull out. In this subcase, the number of lanes more than two lanes in the same direction of travel of the road is not an important parameter, as the interaction will primarily occur in the rightmost lane, i.e., testing on a road with 2, 3, 4, or 5 lanes in the same direction of travel likely will have the same outcome. Surrounding traffic may play an important role in this subcase, so the presence of parked vehicles on the side of the road occluding the view to the pulling out vehicle and the escape paths available to the ADS in adjacent lanes may affect performance and should be accounted for in variations of the scenarios. Parameter ranges, such as vehicle speeds, accelerations, and times to collisions, are varied based on the test. As discussed previously, data from human crash or naturalistic data sources or previously collected data from on-road testing will be used at this stage to determine whether the conflicts are representative of the types of scenarios the ADS might reasonably encounter in a given ODD. The result of this step in the concrete scenario creation process is a set of test specifications, including the scenario, relevant salient factors, and parameters to test. Test engineers use this test specification to create concrete scenarios, either by collecting data from a test track or creating virtual scenarios according to the specification. We are using the stratification of scenario definitions of functional, logical, and concrete first introduced in the PEGASUS project (Menzel et al., 2019) and since adopted in other standards (e.g., ISO 34502 [ISO, 2022], the UN ECE VMAD [UN ECE, 2022]) where a concrete scenario "specif[ies] a concrete value for each parameter and, thus, are the basis for reproducible test cases" (Menzel et al., 2019).

The test execution phase is tracked using an issue management system that provides traceability and review for each stage. The scenario category combinations considered are contained within safety groups. Each set of concrete scenarios generated is tracked by a test request that is entered into the issue management system as a child of the scenario category combinations. The test request entry will include the specification of the concrete scenario, which both allows the test engineers to generate the scenarios and a review mechanism for those responsible for the evaluation to audit and approve the decomposition of high level scenario category combinations into concrete scenarios. Individual concrete scenarios may be only applicable to certain ADS platforms based on ODD and the issue management system is used to track which concrete scenarios make up the final evaluation set for a given program. In addition to tracking that scenarios are identified and created in a traceable manner, the virtual test platform includes tooling to repeatedly build and execute the software under test and simulator environment. Once scenarios are evaluated in the virtual test platform, the results are stored in a database that can be used for analysis. Many of the results and metrics are computed automatically and displayed to users as dashboards that aid in interpretation of the results.

The collection of these representative scenarios use variational techniques to increase the coverage of the created scenarios. Test requests will define logical scenarios that include ranges of parameter values (such as speeds, accelerations) where multiple concrete scenarios will be collected with different parameter values. Additionally, the same concrete scenario with the same parameters may be executed multiple times, which can result in slight parameter variations due to testing equipment, simulation, measurement, and/or repeatability.

## Safety Evaluation

This section describes the safety evaluation that is done leading up to the decision on readiness to start fully autonomous operations in a given ODD. The CAT methodology is executed and a report is generated for the results of the analysis.

The primary result is measurement of the ADS performance against the acceptance criteria laid out in the safety test objectives. After the set of concrete scenarios for a given ODD are identified and created, each scenario is evaluated using the virtual test platform for the candidate release software. A candidate release software version is identified through Waymo's software control and release process. Next, the software is evaluated using the SiL platform discussed previously on the set of identified concrete scenarios. Each simulated scenario is evaluated for the collision and serious injury metric, and the results are aggregated to assess the software candidate's ability to meet the acceptance criteria.

In addition to primary evaluation of the acceptance criteria, several analyses are performed to aid interpretation of the results. First, a coverage assessment is presented that contextualizes how this ODD differs from past ODDs, and how the considered scenarios have changed (i.e., carried over from prior releases or newly created based on a ODD change). Second, notable results are highlighted, including strengths and areas for improvement based on performance regressions and/or expansion of the ODD compared with previous ODDs. The interpretation of the results can be used to compare results of other readiness determination methods to determine if there are trends that might modify the scope or scale of fully autonomous operations. The reproducibility analysis described in a previous section (where the simulated results are compared to test track results using the same software) is also performed for every start of fully autonomous operations.

# Results

Waymo has operated a public, fully autonomous ride hailing service, Waymo One, since 2020 that services the East Valley of the Phoenix, Arizona metropolitan area (parts of Chandler, Mesa, and Tempe). This ODD includes 24 hours per day, 7 days a week driving on roads up to a speed limit of 45 mph. Waymo also has fully autonomous ride hailing operations that service parts of San Francisco, California and downtown Phoenix, Arizona (not the East Valley area) using the same hardware and software configurations. As an illustration of the flexibility of the CAT methodology, this results section will detail how the CAT methodology was executed for the start of the fully autonomous ride-hailing operations in San Francisco and Phoenix. In the rest of the results section, we will refer to the initial Waymo One service in the East Valley of the Phoenix area as the CHD (for "Chandler") fully autonomous operations, and the San Francisco and Phoenix ride-hailing operations as the SF/PHX fully autonomous operations. The results in this section are provided as examples of the types of scenarios considered when starting fully autonomous operations of an ADS in an ODD. As the CAT process is performed continuously as the ODD and system are developed, the examples in this section are not comprehensive.

Test Scenarios

*Overview of Scenario Identification*

As Waymo has existing ride hailing operations for which the CAT methodology was applied, the starting point for identifying hazardous conditions for the SF/PHX operations was the scenario database from the CHD operations. Table 1 summarizes differences in ODD between the operations. This table is not an exhaustive list of elements, but instead focuses on differences between the ODDs. Compared to the CHD operations, the SF/PHX operations are on lower speed limit roads and do not include roundabouts. This means scenarios specific to these ODD elements can be excluded. The SF/PHX ODD has some unique ODD elements, such as crossing light rail tracks at intersections, that were not applicable to the CHD ODD.

**Table 1.** ODD Comparison between CHD and SF/PHX Fully Autonomous Ride-hailing Operations, focusing on differences.

| **ODD Dimension** | **CHD Fully Autonomous Operations** | **SF/PHX Fully Autonomous Operations** |
|---|---|---|
| Land Use | Urban/Suburban | Dense Urban |
| Maximum Speed Limit (inclusive) | 45 mph | 30 mph |

| Tunnels | Yes | No |
|---|---|---|
| Crossing Light Rail Tracks at All-way Stops | No (N/A) | Yes |
| Crossing Light Rail Tracks at Traffic Lights | No (N/A) | Yes |
| Roundabouts | Yes | No |

Although the characteristics of the ODD provide some high-level differences that would impact which scenarios are considered, it is difficult to understand intuitively what novel hazardous situations might occur in a dense urban environment like SF/PHX compared to a more suburban environment in CHD. In order to build the scenario database for the SF/PHX ODD, the process shown in Figure 4 was followed. Based on the product requirements and differences in the ODD, a set of core scenarios was extracted from the CHD scenario database. These are common scenarios that drivers will encounter, whether driving in suburban or urban areas. For example, pedestrians appearing from behind parked vehicles, other vehicles running stop signs or red lights. Based on the differences in the driving environment, however, key areas where new scenarios not considered previously could occur or occur at higher rates in SF/PHX than the CHD ODD were identified. Based on these categories, driving data from instrumented Waymo vehicles operated by autonomous specialists collected in the SF/PHX ODD was analyzed to quantify different observed behaviors. Because the types of responder role conflicts are exceedingly rare during data collection, the driving data was analyzed to find situations that may lead to conflicts. For example, the density of pedestrians was measured as a proxy for the likelihood of a conflict with a group of pedestrians crossing the street. These behaviors were compared to the carried over core scenarios, and if there was a lack of coverage then new scenarios were added. The combination of these core scenarios and the newly created scenarios made up the scenario database for the SF/PHX ODD.

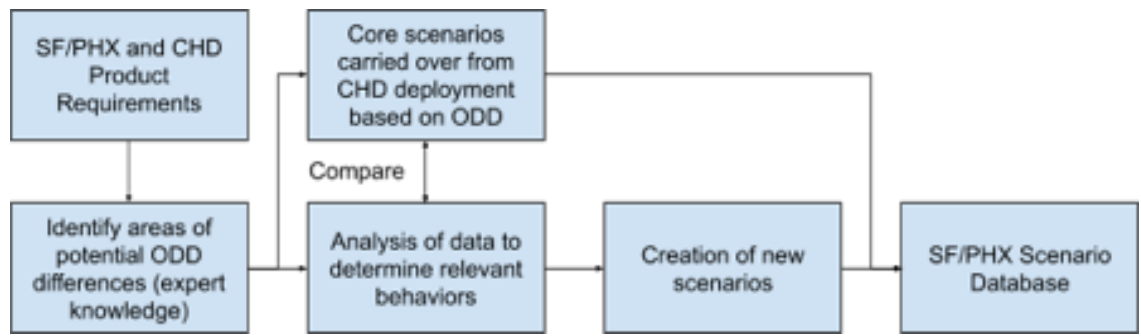

**Figure 4.** Process flow for Creating SF/PHX ODD Scenario Database by Combining Core Scenarios from Previous CHD Scenario Database and Creating New Scenarios.

During the identification of potential ODD differences, one of the largest differences in a dense urban environment such as SF/PHX compared to a suburban area like CHD is the population density which results in more objects located in a smaller area. This increased density leads to higher exposure to and potentially novel scenarios involving parked and double parked vehicles, vehicles pulling out from various street parking configurations, and scenarios involving transportation pick-up and drop-offs. The population density also leads to increased interactions with pedestrians. Analysis of Waymo's data collection prior to the start of autonomous operations showed that there were between 8 and 9 times more pedestrians in the SF/PHX ODD than the CHD ODD. To illustrate the process of identifying new scenarios, the following section presents some examples of novel scenarios involving pedestrians that are applicable to the SF/PHX ODD.

## Example Analysis of Expanded Coverage - Pedestrians

### Pedestrian Crowds

One consequence of higher population densities is the higher possibility of crowds of pedestrians. An analysis of data collected by Waymo vehicles found that areas with more than 20 pedestrians in a 10 second window were 50 times more likely in SF than the suburban environment of CHD. Figure 5 shows two examples of crowds of pedestrians observed in SF. These types of scenes could be challenging for an ADS because multiple pedestrians need to be tracked separately and predictions about their movements need to be made. If a pedestrian suddenly breaks from a crowd into the path of the ADS, the ADS needs to react quickly. Based on this type of scene and challenge, we created a series of scenarios where pedestrians break out from a group of pedestrians into the path of the ADS. An example of this type of scene is shown in Figure 6. The scenarios in this category were varied based on the location on the road (midblock crossing like shown in the figures below, as well as at intersections in a crosswalk), the maneuver of the ADS (going straight or right/left turning onto another road), and the type of pedestrian crossing (single pedestrian crossing like shown below, or a group crossing).

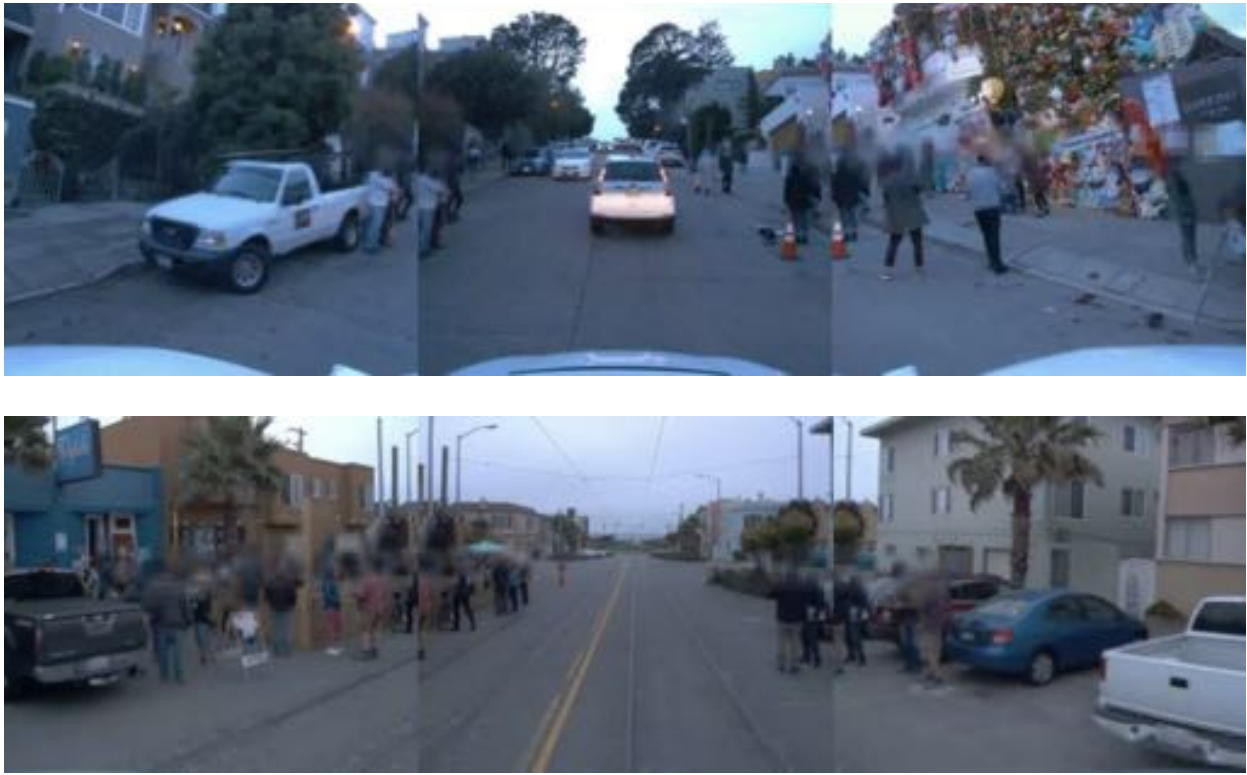

**Figure 5.** Crowds of Pedestrians Gathering to Observe Holiday Decorations (top) and Gather on the Street Outside a Restaurant (bottom)

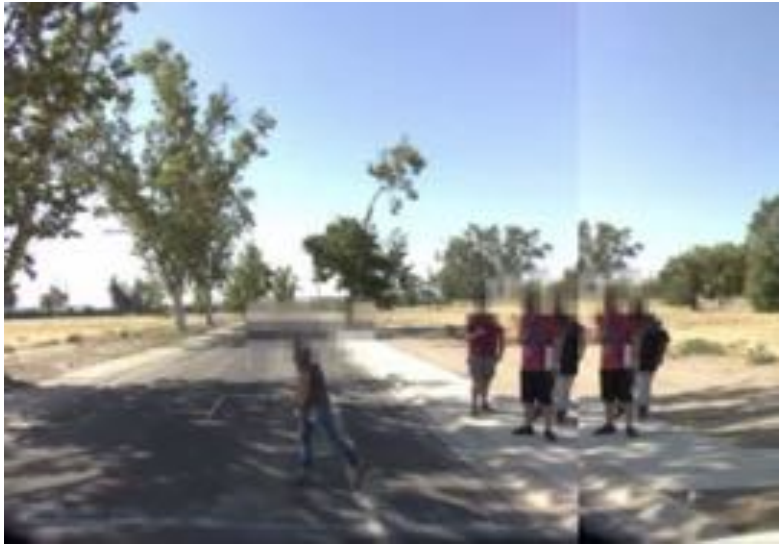

**Figure 6.** Example of a Scenario Created on a Closed Course where a Pedestrian Breaks out from a Group into the Path of the ADS.

Pedestrians on Scooters and Skateboards

Pedestrians that ride on scooters and skateboards also present challenges to an ADS. These scooter and skateboarders have an appearance similar to an ambulatory pedestrian, but can travel at much higher speeds and have sudden changes in direction compared to these ambulatory pedestrians. As a result, it is important for the ADS to differentiate these scooter and skateboard riders from ambulatory pedestrians, and accurately predict their speed and intentions in order to react to collision avoidance situations. Like pedestrian groups, scooter and skateboard riders are encountered 8 to 9 times more frequently in SF and DT PHX compared to CHD. Furthermore, scooters and skateboard riders traveling close (< 3 m) and fast (> 5 m/s) occur at a rate 17 times more frequently in SF and DT PHX compared to CHD. The core scenarios from CHD had sufficient coverage of scooters and skateboarders either (a) traveling on the road like vehicles (e.g., straight crossing path or turn across path conflicts) and (b) traveling in crosswalks like pedestrians. One area that was not well covered by the core scenarios was when a scooter or skateboard rider would either start in a crosswalk and then suddenly exit the crosswalk or when a scooter or skateboarder travels in the wrong direction on the road. Figure 7 shows an example of the former scene. Note that this scenario was made more challenging by parking a pickup truck (see in the left of Figure 7) close to the intersection to occlude the view of the scooter to the ADS.

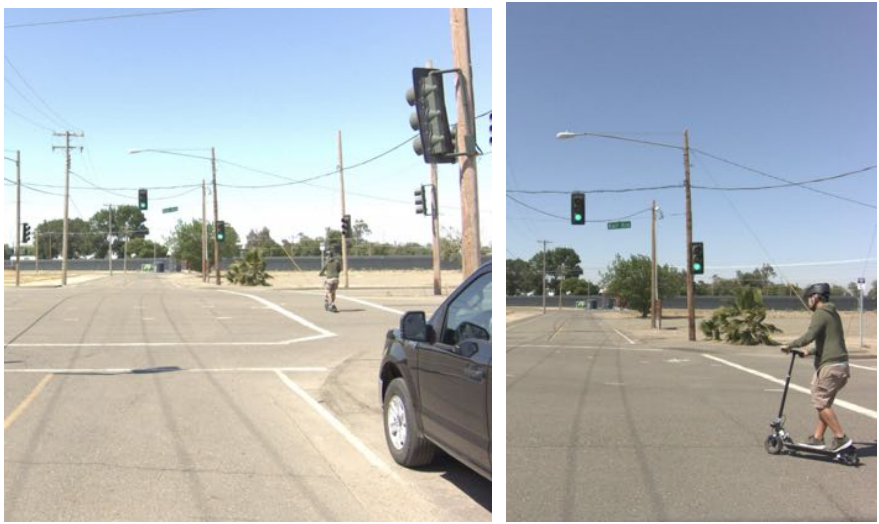

**Figure 7.** An Example Scenario of a Scooter Rider Crossing in a Cross Walk (left) then Suddenly Entering the Path of the ADS (right).

Pedestrians Interacting with Parked Vehicles

An analysis of map data found that SF and PHX has 34% more street parking than the CHD area. In addition, double parked vehicles are more common in SF and PHX compared to CHD. Both of these types of increased vehicle density means there is more opportunity for pedestrians to interact with parked vehicles in SF/PHX compared to CHD. Additionally, delivery trucks that are double parked in the street are more common in SF/PHX than CHD. Occlusions caused by parked or double parked vehicles is a potentially challenging collision avoidance scenario. Although the core scenarios had pedestrians appearing from behind parked vehicles, there were not as many scenarios including double parked vehicles, i.e., stopped vehicles in a travel lane as opposed to parked in a parking lane off the road. Additional scenarios with pedestrians appearing from double parked vehicles or from behind large objects in travel lanes were added. One potentially challenging scene that was added to the scenario database was a pedestrian jumping from the back of a box truck into the path of a vehicle, as shown in Figure 8.

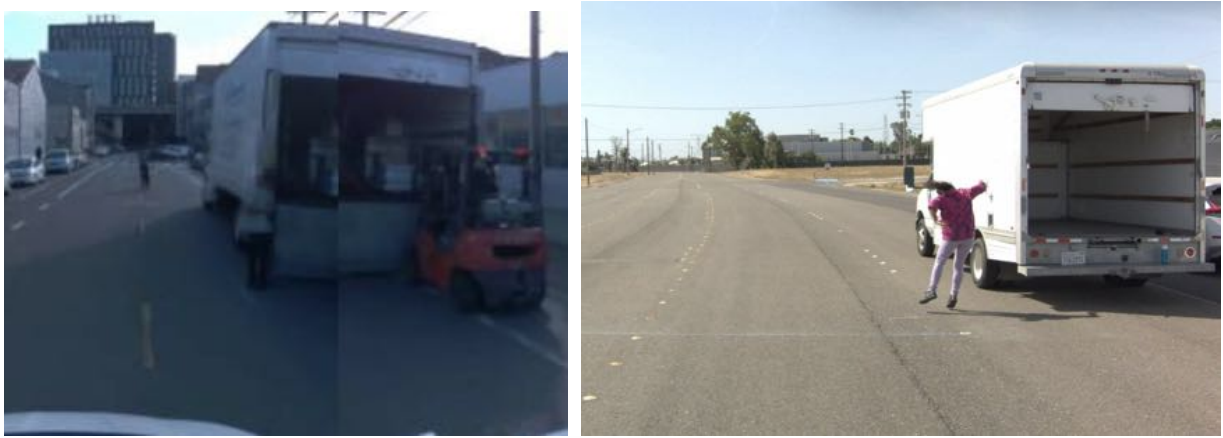

**Figure 8.** An Example of a Double Parked Delivery Truck in San Francisco (left) and a Scenario with a Pedestrian Jumping from the back of a Double Parked Delivery Truck (right)

Safety Evaluation

In total, the CAT evaluation for the SF/PHX ODD contained over 13,000 scenarios. Of these, 48% of scenarios involved a potential conflict with a pedestrian or cyclist and the other 52% involved a potential conflict with a vehicle. There were two road user groups (Vehicles and VRU) and 35 safety groups where the aggregate metric evaluation was performed. The result of the safety evaluation was that the ADS met the acceptance criteria based on the comparison to the reference model for both collisions and serious injury events.

As presented in the previous section, the primary method for determining relevant scenarios was analysis of Waymo driving data and a systematic identification of relevant scenarios. On-road testing data leading up to the start of fully autonomous operations, however, is a useful confirmation that the identified representative scenarios cover the challenges that are observed during testing done with an autonomous specialist (test driver) behind the wheel, before fully autonomous operations. To accomplish this validation, a set of simulated collision events that occurred after an autonomous specialist disengaged the system were reviewed. Of the 419 simulated events reviewed that featured the ADS in the responder role, 106 (25%) were not covered in the existing scenario database and prompted additional scenarios to be added. Of the events not covered by the existing scenarios, 46 (43% of the not covered events) involved vehicles leaving parking spots or driveways in novel ways. This result shows that most responder role scenarios experienced during on-road testing were already covered, but a feedback loop with on-road testing is an important mechanism to help ensure coverage of representative situations and reduction of the unknown-unknown scenario space.

In addition to the results of the primary safety evaluation (i.e., collision and serious injury evaluation), the results of the CAT evaluation are compared to prior releases and to the performance of other readiness determination methods to attempt to determine if there are any trends that may not be apparent in the primary safety evaluation. For example, an analysis comparing the ADS's performance to the previous release was performed and found that the ADS performance in one pedestrian crossing in front of the vehicle safety group had lowered, although it was still meeting the acceptance criteria. Upon further examination of these scenarios, the cause in the performance change was a change in behavior of the ADS, where it was braking harder and earlier than in the previous release in which the ADS applied only light braking and passed closely by the pedestrian. As a result of this earlier braking, there were more absolute number of simulated collisions than in the previous software release. When compared to the reference behavior model, however, there was only 1 more simulated collision involving both the ADS and reference model as this changed behavior of the ADS to slow instead of passing also affected when the reference model started to perform evasive maneuvers. Although by the measure of contacts in the simulation there was a regression in performance, the behavior in the proposed release was more conservative (i.e., braking to slow down) than the previous release (passing by closely). Because of this more conservative behavior and that the acceptance criteria were still met, there was no further testing or changes made to the testing program. This detailed examination, however, provides more opportunities to discover behavior that can be addressed and improved upon prior to a release. This type of analysis and review of performance results is an additional check that ADS meets a level of acceptable risk across all expected responder role scenarios, without compromising performance in one area for performance in another. This example also points to challenges in constructing these scenarios and assessing performance differences across multiple software versions.

## Discussion

The results of the CAT methodology used for the start of fully autonomous ride-hailing operations in San Francisco and Phoenix was presented here as an example of implementation of this process, but the methodology has been used in previous Waymo operations such as the ODD of the publically available Waymo One service around Chandler, Arizona, and is being applied to other ADS applications, such as trucking. Across these different platforms and ODDs, the methodology for identifying hazardous situations is similar. Different platforms and ODDs, however, may require slightly different approaches in test case allocation. The SF/PHX program described in this paper heavily utilized simulations based on data collected on test tracks, and derived from multiple data sets, such as human and ADS driving databases. For a heavy vehicle, highway ODD program, the vehicle masses and speeds are much higher than the ride hailing ODD described in this paper. Although surrogate vehicles (e.g., guided soft targets) can be used to reduce the risk to equipment and test track personnel; those platforms have dynamic constraints (e.g., lateral velocity and top speed) that may make it impossible to collect all required scenarios on a test track. Due to these constraints, a heavy vehicle, highway ODD program may require more reliance on fully synthetic scenarios compared to ride hailing applications. In our experience, the performance of an ADS in a full system evaluation such as that done in the CAT methodology is also highly dependent on the fidelity of the sensor data. A heavy reliance on synthetic scenarios should be accompanied by validated realistic sensor and perception simulation. Even in the approximately 6 years that Waymo has been developing the CAT methodology, simulation capability has significantly increased its ability to synthetically generate realistic sensor data, supporting our decision to develop a greater share of the newly generated concrete scenarios using simulated sensor data today than in the past.

The acceptance criteria used in the CAT methodology that is based on a comparison to a reference behavior model is similar to other proposed approaches. For example, Rothoff et al. (2019) performed a safety impact assessment on a fictive ADS and compared its performance to both the humans involved in the original collisions and a reference model representing an attentive, skilled driver. Although the CAT methodology does not attempt to perform a safety impact assessment (i.e., predict the expected number of collisions or serious injury events), using a reference behavior model is well supported by previous research and industry standards, such as ISO 34502:2022 (ISO, 2022b) and ISO/TR 21934-1:2021 (ISO, 2021).

As discussed in the test case allocation section, the CAT methodology implements a segmentation approach where scenarios representative of a hazardous situation are sampled instead of all possible scenarios being exhaustively sampled and evaluated. The alternative, exhaustive sampling of scenarios, presents difficulties in being able to effectively evaluate performance. For example, the perception performance may differ based on the properties of an object (e.g., varying vehicle configurations or object shapes or types). If all possible appearances of objects were tested in the CAT methodology, the number of scenarios evaluated may need to be hundreds to thousands of times larger. If other dimensions are added, such as evaluating at all increments of speed within a range, the testing program quickly becomes intractable. The impracticality of such an approach would be a substantial impediment to further development of ADS technology without any demonstrable benefit. Because this type of exhaustive sampling is not performed, however, there is additional evaluation. The perception subsystem, for example, is evaluated on its ability to perceive various actor types during the sensing system validation (Webb et al., 2020). This evaluation and its associated requirements on perception accuracy give us confidence that the CAT methodology performance across different variations of the same object is equivalent. Lighting and visibility conditions (e.g., low sun angle, fog) can also affect perception performance, especially the camera and LIDAR systems used in an ADS. As these types of conditions are added to the ODD, the CAT scenario database can be evaluated using synthetic modifications to the perception data in the simulation. For example, for the SF/PHX ODD the entire CAT evaluation was performed again for simulated light fog as this condition is within the ODD.

An alternative to the segmentation approach used in the CAT methodology are various adaptive sampling methods, where the performance of the ADS under test is used to select new concrete scenarios that may have a higher likelihood of an adverse outcome. This adaptive approach is useful for finding novel situations where the ADS may fail, often called expanding the "unknown-unknown" scenario space. In addition to the parameter explosion issue discussed above, another challenge with adaptive testing is that it is difficult to compare performance from software release to software release if the underlying scenarios are different. Having a finite set of scenarios allows these release-to-release comparisons. The Waymo CAT evaluation can also be used to test individual software changes during the software development process before a release candidate is identified. For these reasons, we believe this approach used in the Waymo CAT methodology is appropriate for the given safety test objective to evaluate the full system performance in urgent responder role situations and in the context of other complementary readiness methodologies (Webb et al., 2020) that address some of the advantages of using an adaptive sampling method.

Through the scenario identification process, the CAT methodology aims to provide a representative sample of responder role conflicts that the ADS might reasonably encounter in its ODD. The unknown-unknown space is explored in the CAT methodology through the scenario collection process, where parameters are varied across a range to find representative scenarios with regard to the reference model performance. Waymo utilizes multiple methods in addition to the CAT methodology to search the unknown-unknown space. One method is through simulated deployments, where historically collected driving data is simulated using new software versions (Webb et al., 2020). This simulation approach allows for a safety evaluation based on much larger volumes of driving miles compared to what is possible by driving test vehicles using a software candidate and leads to the discovery of previously unknown failures or possible regressions (situations that were handled correctly in a previous software release, but not in the new candidate). Waymo also uses other types of synthetic simulation techniques to generate challenging scenes that may expose unknown-unknown failure modes (Igl et al., 2022; Waymo, 2021). Situations that are found through simulated deployments, on-road testing, or other means are also reviewed and used to expand the considered scenarios in the CAT scenario database, such as the review of autonomous specialist disengagement simulations described in the results section.

The metrics used to assess the acceptance criteria in the CAT methodology are based on comparing the ADS to the NIEON model on the number of collisions of any severity and those that have a risk of serious injury (i.e., probability of MAIS3+ injury). Although much of the past safety impact methodology research discussed in the introduction section has considered serious injury and fatality collisions, the hope is that by introducing ADS that the risk of serious injury and fatality collisions would become low. If so, collisions that result in lower levels of injury, for example, MAIS1 or MAIS2, should also become a focus of safety evaluations. By considering separately (a) the overall collisions of any severity and (b) serious injury collisions, the metrics presented in this paper do implicitly consider these AIS1 and AIS2 level injuries. There are some challenges associated with adding, for example, a MAIS1 and/or MAIS2 metric to the CAT methodology. Namely, because many injury risk curves are developed based on retrospective crash investigations (see for example, McMurry et al., 2021), data on AIS1 and AIS2 injuries are more limited in number and quality than data sources that have AIS3+ injuries. As more research is done in quantifying injury risk functions for AIS1 and AIS2 injuries for all road users (vehicles and VRUs), the CAT methodology could be adapted to also consider metrics related to these lower levels of injury.

The response time model used in the NIEON reference behavior model that is described in Engström et al. (2022) is fit on human naturalistic collision avoidance behavior. The data presented in Engström et al. (2022) is from passenger vehicle drivers. In order to expand this model to other vehicle types, for example, heavy trucks, additional analyses would need to be performed to ensure the model presented in Engström et al. (2022) is applicable to drivers of these different vehicle types. In addition, the collision avoidance maneuvers (i.e., steering and braking) as was presented in Scanlon et al. (2022) may also differ by vehicle type and would need different parameters based on naturalistic driving data.

The CAT methodology is a scenario-based testing approach that is used in conjunction with Waymo's other safety methodologies to demonstrate that the Waymo ADS does not present an unreasonable risk to motor vehicle safety in responder role scenarios under nominal operating conditions. The scenario identification process starts before ADS software is ever tested in the real world by identifying hazardous situations through a structured process and by using available historical crash and near-crash data, from both human drivers and ADSs. Over time, as new information becomes available through internal development or through new external data sources, the set of scenarios is confirmed to be representative of an ODD, or missing scenarios are added based on this new information. The result is that by time when the safety evaluation is performed prior to starting fully autonomous operations, we have a high degree of confidence that the scenario database contains a representative sample of responder role conflicts that the ADS might reasonably encounter in its ODD. Furthermore, the CAT methodology that is described in this paper is the approach taken at the time of writing. The methodology has and will continue to evolve and improve over time.

## Conclusions

This paper outlines an approach for and provides results of a large-scale scenario-based testing program for safety assessment of a SAE level 4 ADS with regard to its capabilities in crash avoidance situations that require urgent evasive maneuvers to avoid a collision. The CAT methodology described in this paper is designed to be a flexible, iterative process that can be scaled across different vehicle platforms and ODDs. The safety objectives are to demonstrate the ADS is able to avoid collisions and mitigate serious injury outcomes with comparable ability as a reference behavior model for collision avoidance behavior in the responder role. Test scenarios are identified in a structured manner, based on a number of data sources, including human and ADS crash and near-crash data, previous ADS testing results, expert knowledge, and experience from on-road testing. Tests are executed in a virtual test platform, either using simulations based on collected sensor data or in purely synthetic simulations. The CAT methodology was applied to an application of fully autonomous ride-hailing operations in San Francisco, California and Phoenix, Arizona, resulting in a set of over 13,000 scenarios. The CAT methodology is one of many methodologies used to inform readiness determinations performed by Waymo prior to deploying a fully autonomous system and is an important demonstration of competency in avoiding urgent, responder role situations. The assessment done for the CAT methodology does not consider ADS failure situations (e.g., sensor failures). These failure situations are evaluated and mitigated through other methodologies such as hazard analysis, platform and perception verification and validation (V&V), and simulated deployments (Webb et al. 2020). This process converges to a set of scenarios that are a representative sample of responder role conflicts that the ADS might reasonably encounter in its ODD. Other methodologies, such as simulated deployments, hazard analysis, and V&V serve as feedback mechanisms to identify additional uncovered scenarios that reduce the "unknown-unknown" hazard space. Combined with a virtual test platform that has been calibrated to the data collected as part of many years of ADS development and testing, results in a robust and scalable safety evaluation for the ADS's behavior in urgent, responder role conflicts.